\documentclass{article}
\usepackage{PRIMEarxiv}
\usepackage[utf8]{inputenc} 
\usepackage[T1]{fontenc}    
\usepackage{url}            
\usepackage{booktabs}       
\usepackage{amsfonts}       
\usepackage{nicefrac}       
\usepackage{microtype}      
\usepackage{lipsum}
\usepackage{fancyhdr}       
\usepackage{graphicx}       
\usepackage{listings}
\usepackage{mdframed}
\usepackage[round]{natbib}
\usepackage{comment}
\usepackage{tabularx}
\usepackage{caption}
\usepackage{float}
\usepackage{xcolor}

\usepackage{bm}
\usepackage{tikz} 
\usepackage{pifont} 
\usepackage{amssymb}
\usepackage{amsmath,upgreek}
\usepackage{titlesec}
\usepackage{subcaption}
\usepackage[title]{appendix}
\usepackage{stfloats}
\usepackage{multicol,subeqnarray}
\usepackage[most]{tcolorbox}
\definecolor{block-gray}{gray}{0.85}

\definecolor{xlinkcolor}{cmyk}{1,0.6,0,0}
\usepackage[bookmarks=false,         
     pdfnewwindow=true,      
     colorlinks=true,    
     linkcolor=xlinkcolor,     
     citecolor=xlinkcolor,     
     filecolor=xlinkcolor,  
     urlcolor=xlinkcolor,      
final=true
]{hyperref}
\graphicspath{{img/}}     
\usepackage{booktabs} 
\usepackage{array} 
\newfloat{listing}{htbp}{loa}  
\floatname{listing}{Listing} 

\title{Injecting New Knowledge into Large Language Models via Supervised Fine-Tuning}

\author{
   \\
  \textbf{Microsoft}\\
Nick Mecklenburg, 
Yiyou Lin, 
Xiaoxiao Li, 
Daniel Holstein,  
Leonardo Nunes, \\
Sara Malvar,
Bruno Silva, 
Ranveer Chandra,
Vijay Aski,
Pavan Kumar Reddy Yannam,
Tolga Aktas,
Todd Hendry
}

\begin{document}
\maketitle

\begin{abstract}
    In recent years, Large Language Models (LLMs) have shown remarkable performance in generating human-like text, proving to be a valuable asset across various applications. However, adapting these models to incorporate new, out-of-domain knowledge remains a challenge, particularly for facts and events that occur after the model's knowledge cutoff date. This paper investigates the effectiveness of Supervised Fine-Tuning (SFT) as a method for knowledge injection in LLMs, specifically focusing on the domain of recent sporting events. We compare different dataset generation strategies – token-based and fact-based scaling – to create training data that helps the model learn new information. Our experiments on GPT-4 \citep{openai2024gpt4} demonstrate that while token-based scaling can lead to improvements in Q\&A accuracy, it may not provide uniform coverage of new knowledge. Fact-based scaling, on the other hand, offers a more systematic approach to ensure even coverage across all facts. We present a novel dataset generation process that leads to more effective knowledge ingestion through SFT, and our results show considerable performance improvements in Q\&A tasks related to out-of-domain knowledge. This study contributes to the understanding of domain adaptation for LLMs and highlights the potential of SFT in enhancing the factuality of LLM responses in specific knowledge domains.
\end{abstract}


\section{Introduction}
In the explosion of generative AI in the last couple years, more and more use cases are lending themselves towards large language model (LLM) applicability. Typically an LLM application developer is faced with the question of how to best adapt some particular model for their downstream task; this question usually involves distinguishing among the techniques of few-shot learning or prompt engineering, retrieval augmented generation (RAG) \citep{NEURIPS2020_6b493230}, supervised fine-tuning (SFT), some variant of reinforcement learning for human feedback (RLHF) \citep{schulman2017proximal, bai2022constitutional}, or perhaps some combination of these techniques \citep{balaguer2024rag}.

The motivation for this paper lies in the recognition that despite the impressive capabilities of pre-trained LLMs, they are inherently constrained by the scope and recency of their training data. Internet-scale corpora, which form the foundation of LLM pre-training, are by nature finite snapshots, limited both temporally and in their breadth of covered knowledge. This limitation poses a significant hurdle for applications that necessitate up-to-date information or domain-specific knowledge that postdates the model's training cut-off or falls outside the scope of the training corpus.

Moreover, the dynamic and continuously evolving landscape of human knowledge further complicates the challenge, as new events unfold and specialized domains generate proprietary or niche content that is not publicly accessible. Consequently, developers aiming to deploy LLMs in scenarios that demand current or specialized knowledge must devise strategies for domain adaptation that effectively incorporate this new information into the model.

While RAG \citep{NEURIPS2020_6b493230} provides an ingenious workaround by augmenting model responses with an external knowledge base, this approach sidesteps the core issue of assimilating new knowledge directly into the model itself. Thus, we are compelled to explore alternative methods that enable LLMs to internalize and retain new information through direct training interventions.

Herein lies the crux of our inquiry: how can we construct a training dataset from a body of documents that facilitates the learning of new knowledge through simple SFT techniques? Addressing this question is not only of theoretical interest but also carries significant practical implications for the deployment of LLMs in real-world settings where accuracy, currency, and domain specificity are paramount.

Our investigation into the domain of knowledge ingestion via direct training has yielded several notable contributions:

\begin{enumerate}
\item \textbf{Analysis of Token-based Q\&A Dataset Generation:} We provide a comprehensive evaluation of the standard practice of token-based Q\&A dataset generation. Our findings reveal that this method may not ensure complete or uniform coverage of new knowledge within a document corpus. This observation is critical as it underscores the potential limitations of prevailing dataset preparation strategies and highlights the need for more targeted approaches to knowledge incorporation.

\item \textbf{Development of a Fact-based Generation Process:} In response to the identified shortcomings, we propose a fact-based generation process. This methodology prioritizes the even distribution of attention across all salient facts within a source document, ensuring that each piece of information is adequately represented in the training data. This approach stands to significantly enhance the model's ability to internalize diverse and detailed knowledge from domain-specific corpora.  

\item \textbf{Empirical Validation of SFT for New Knowledge Learning:} We demonstrate that even straightforward SFT can lead to substantial improvements in model performance when handling out-of-domain, post-cutoff knowledge. Our results not only validate the effectiveness of our proposed fact-based generation process but also provide a compelling case for the practicality of SFT as a tool for domain adaptation in LLMs. This contribution has profound implications for the application of generative AI in dynamic fields where staying abreast of the latest information is crucial.  
  
\item \textbf{Benchmarking against Retrieval-Augmented Models:} We extend our study to include a benchmark comparison between our SFT models and those employing RAG. This analysis provides insights into the trade-offs and relative merits of direct training versus retrieval-based augmentation, offering valuable guidance for practitioners in the selection and implementation of knowledge ingestion methodologies.  
  
\item \textbf{Exploration of Hyperparameter Sensitivity:} Recognizing that the tuning of hyperparameters is often a nuanced and impactful aspect of model training, we delve into the sensitivity of our models to a few hyperparameter settings. Our exploration sheds light on the robustness of our findings and sets the stage for future work on optimization strategies tailored to knowledge ingestion tasks.  
\end{enumerate}

Here, we illustrate not only the breadth of our contributions but also their potential implications for the field of AI and LLM application development. By doing so, we aim to provide a thorough understanding of the impact of our research and its relevance to both academic inquiry and practical application.
\section{Related Work}
There are several bodies of related work on the topic of post-training domain adaptation for language models, with emphasis on factuality. One such body is the study of knowledge injection, which sets out with the same goal of encoding new knowledge into language model systems \citep{wang2020kadapter, Chen_2022}. Past popular knowledge injection methods might involve retrieval/augmentation using some external knowledge base \citep{fan2020enhanced, song2016better}; the knowledge base need not be a vector database as is state-of-practice in modern RAG systems \citep{NEURIPS2020_6b493230}, instead falling to alternative constructions like knowledge graphs \citep{martino2023knowledge}. Another common formulation of knowledge injection involves using model adapters, such as \cite{lauscher2020common}, where only the adapter parameters meant to encode the domain adaptation are left unfrozen during training in an attempt to mitigate catastrophic forgetting.

Catastrophic forgetting, or the tendency of models to freely overwrite structure learned for any previous task in the learning of some subsequent one \citep{MCCLOSKEY1989109}, is itself of particular import for our objective. Avoiding such forgetting partially motivates the complexities of external augmentation techniques or partially-frozen adapter training.

This is not to say direct training approaches are avoided altogether, however. \cite{xu-etal-2023-kilm} tries continued pretraining using a knowledge infill objective to boost factual retention. \cite{tian2023finetuning} uses the constrained optimization of direct preference optimization \citep{rafailov2023direct} to encourage factuality during post-training. We take particular interest in \cite{ovadia2024finetuning}, which leverages unsupervised training on chunks of post-cutoff Wikipedia articles, paraphrased some number of times, to teach the model new knowledge. Their approach of using careful repetition to boost FT performance matches our choice in this study, though we differ in our techniques of supervised vs. unsupervised learning, where they choose unsupervised to effectively do continued pretraining for knowledge injection. \cite{ovadia2024finetuning} also does not consider the fact density of the underlying corpuses they set out to learn, aligning more with our token-based scaling as outlined in section \ref{sec:datagen}.
\section{Dataset Generation}
\label{sec:datagen}
We identify six Wikipedia articles, five about recent sporting championships -- 2023 Cricket World Cup \citep{enwiki:1213361100}, 2023 American Football Superbowl \citep{enwiki:1215541265}, 2023 FIFA Women's World Cup \citep{enwiki:1214162630}, 2023 PGA Championship \citep{enwiki:1213024529}, 2023 NCAA Men's Basketball Division I Tournament \citep{enwiki:1215798602} -- that were post-cutoff and one about a pre-cutoff sports event, the 2018 FIFA World Cup \citep{enwiki:1216022372}. Sporting events were chosen as they are fact/statistic rich, easy to understand, and related information is mostly black-and-white for being pre- or post-cutoff. For each of the documents, we use the text extracts API with explaintext enabled to fetch the plain text portions of each section, returned in JSON format. We filter and clean the sections, removing any blank ones or ones without meaningful plain text contribution (such as those that are entirely table-based save for some annotations).

For each of our cleaned documents, we generate two types of datasets: token-based scaling and fact-based scaling. The motivation for having two sets of datasets is detailed in section \ref{sec:tokenfactcov}. We endeavor to scale Q\&A datasets with different questions covering the same content -- as opposed to simply training on the same data for more steps -- as the increased examples will allow the model to see the contained target knowledge in more ways. This varied repetition increases the effectiveness of learning while mitigating overfitting as much as possible that would otherwise lead to significant performance degradation \citep{ovadia2024finetuning}. We do not want the model to memorize the phrasing of the source document, but rather the knowledge contained within it.

\subsection{Token-based}
Our steps for generating the token-based scaling datasets are:
\begin{itemize}
    \item Initialize an empty question bank and seed it with one Q\&A pair written manually based on the article's overview section.
    \item For each section in the document, calculate its token count using OpenAI's tiktoken library. Repeatedly query base GPT-4 with temperature 1 and top\_p=0.95 (which we use for all dataset generation tasks in this section) to generate question-and-answer pairs until the generated token count between unique questions and answers for that section surpasses 10 times our source section tokens.
    \item We are able to create our per-section 1x and 5x datasets using a subset of the 10x generations. All questions are unique from an exact match standpoint.
    \item The 1x, 5x, and 10x fine-tuning datasets are created from the union of the per-section ones for each respective scaling factor.
\end{itemize}

The evaluation sets for each document were generated according to an identical procedure, to 1x token scale. All questions were again ensured to be unique and not contained within the train set already.

\subsection{Fact-based}
For fact-based scaling, first we needed to generate the list of atomic facts contained in the documents. This was done by querying GPT-4 with the cleaned document sections from before. Then, for each document:
\begin{itemize}
    \item We iterate over the atomic facts and generate 10 unique question-answer pairs by querying GPT-4.
    \item Pairs are only accepted if they are not already present in the question bank so far so as to avoid duplication, else a new pair is generated.
    \item From this 10x set, we create the 1x and 5x scaling subsets.
\end{itemize}

In the fact-based dataset generation prompt, we give GPT-4 the option to skip Q\&A generation if the fact is too broad, unclear, or otherwise not related to the topic of the document. We manually inspect the few such cases where the model elects to follow this option. We either filter out those atomic facts if they are indeed unrelated or, if they are kept, regenerate question-answer pairs with the ``skip" prompt option removed.

An example of one such fact that GPT-4 chose to skip is "Russia is a nation that lost millions of lives in fighting Nazism", which was generated for the 2018 FIFA World Cup article that was hosted in Russia. A statement along these lines was mentioned in a quote in the original document, and the model (correctly) deemed this too unspecific to the sports event to be included. An example fact that GPT-4 skipped which we still generated question-answer pairs for would be for the American football Superbowl LVII article: "In February 2022, over 200 liberal religious leaders petitioned NFL Commissioner Roger Goodell." While seemingly less related, this concerned some calls to move the location of Superbowl LVII from its original Arizona venue.

Fact-based evaluation sets were generated on the accepted set of atomic facts using the prompt without the skip option, deduplicated on an exact match criterion, and at 1x scale.

We include samples of each of the prompts used for dataset generation in appendix \ref{sec:sampleprompts}. Samples of atomic facts identified for each article are listed in appendix \ref{sec:atomicfacts}.
\section{Training Methodology}
The same training settings are used for both token and fact-based datasets for all scaling factors. We choose the model GPT-4 -- v0613, which has a cutoff date of September 2021 \citep{msftaoaimodeldocs} -- for our fine-tuning. Our SFT procedure is parameter efficient, leveraging the lora technique \citep{hu2021lora} with rank 16, batch size of 1, and 3 epochs. The context length is large enough to fit in all of our training examples. Gradient updates are backpropagated only over the assistant tokens, not the user prompt tokens upon which we condition them.

Note we do not exhaustively sweep over all hyperparameter configurations, so the ones chosen here are not optimal. We find especially that there is improvement to be made in the learning rate and epoch count parameters as we observe some signs of underfitting. We discuss the implications of this more in section \ref{sec:limitations}.

We clarify these settings are not necessarily the same hyperparameters / algorithms / settings as those used in the Azure OpenAI or OpenAI finetuning products as we do not use these interfaces.
\section{Evaluation}
After training GPT-4 on each dataset, we end up with a set of lora deltas that we use for inferencing. We inference with a top\_p of 1 and temperature 0. Our fact-based and token-based dataset training runs are evaluated on the fact-based and token-based evaluation sets, respectively. For cross-validation, we evaluate the token-based models on the fact-based evaluation set as well, which we discuss in section \ref{sec:crossval}. The motivations for not checking fact-on-token are mentioned in section \ref{sec:limitations}.

For assessing the correctness of the trained models' answers on the evaluation questions, i.e. retention of the source documents or atomic facts, we again query GPT-4 for a binary assessment. The prompt used for this is also included in appendix \ref{sec:sampleprompts}.
\section{Results}
\subsection{Token-Based Scaling}
\label{sec:tokscale}
\begin{figure}
\centerline{\includegraphics[width=\textwidth, keepaspectratio]{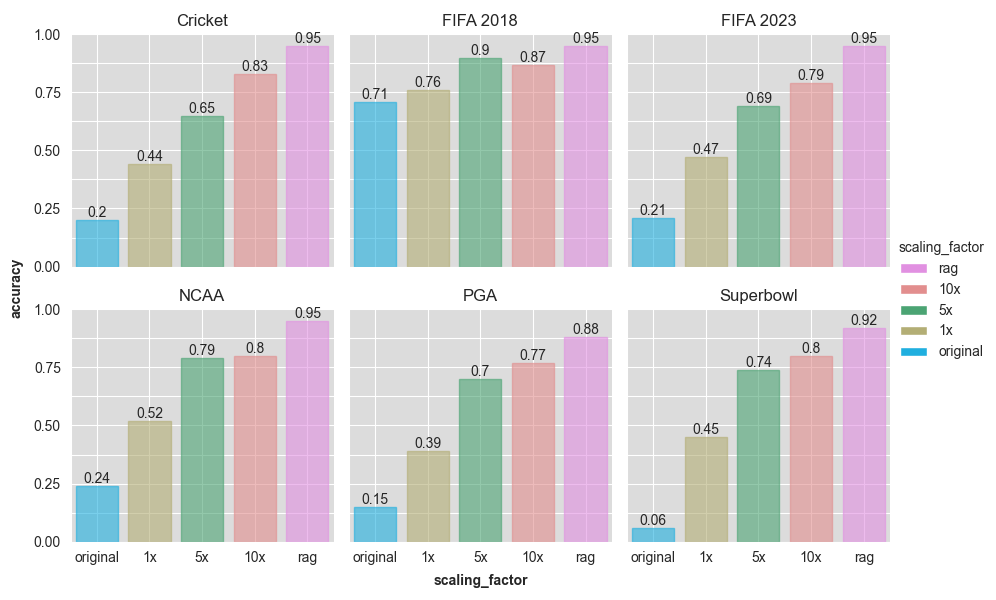}}
\setlength{\belowcaptionskip}{-8pt}
\caption{\textbf{Token-based evaluation set} accuracy for our six documents across 1x, 5x, and 10x scaling with models trained on \textbf{token-scaled datasets}. The base model results with no training are included under the bars annotated as ``original," and we include a RAG baseline as well which leverages the cleaned document sections to answer the eval questions.}
\label{fig:hist}
\end{figure}
The token-based scaling results are shown in figure \ref{fig:hist}. First, we compare the results for the 2018 FIFA World Cup vs. the others. We notice the untrained model for this dataset scores much better than any of the others (0.712 accuracy vs. a max of 0.242), serving as a sanity check that indeed the other five articles are post-cutoff and out-of-domain. The knowledge we try to inject into the model is mostly new.

We observe variation in the base model's performance between the post-cutoff datasets (Superbowl's 0.061 accuracy vs. NCAA's 0.242). This is attributed to some details of the sporting events being decided in advance. For example, the host city and venue selections for the 2023 FIFA Women's World Cup were announced in March 2021 \citep{enwiki:1214162630}, before the cutoff, so it would be fair game for the model to answer questions about those details. The majority of the information from the articles are still post-cutoff though, as noted above when comparing to the FIFA 2018 results.

Between the base model and the 1x train dataset, the accuracy more than doubles for each of our five post-cutoff events. Training on the 5x scaled datasets continues with statistically significant, large gains for all events, with PGA seeing the largest relative improvement at +78.6\% compared to the 1x case. The gains quickly drop off for some datasets with 10x scaling, however, with NCAA seeing a modest 1\% relative improvement. These are diminishing returns. Cricket, though, sees a 26.8\% relative accuracy improvement in 10x compared to 5x, suggesting it still had content not well covered by the 5x dataset that training for 10x helped it improve. 

We see the effects of token scaling on fine-tuning effectiveness is greatly diminished for the FIFA 2018 dataset across the board compared to the other five. This makes sense intuitively, since that 2018 event, being pre-cutoff, has already been trained on by the model. We do see modest relative gains of 17.4\% in the 5x model compared to the 1x one, but performance notably degrades for the 10x case. We suspect this is due to the lack of diversity in question generation (same style/distribution of Q\&A) causing overfitting and a poorer ability to generalize. Across the internet-scale corpus that GPT-4 was pretrained on, details and statistics about the FIFA 2018 World Cup likely appeared in many different dialogues, tones, and perspectives; post-training now on one form with such a high degree of repetition may be ablating some of the generalization benefits we otherwise got from the original training data. 

This lack of diverse question generation as we hit the 10x scale may similarly explain why we see diminishing returns for some of our other datasets that are post-cutoff, and we suspect a similar improvement in our dataset generation methodology may increase training robustness and give us a further performance boost (constrained for same compute/budget, of course).

Since RAG is a go-to solution for new knowledge injection into LLM systems as mentioned previously, we perform RAG over the cleaned document sections using Azure OpenAI hybrid search (vector + keyword) \citep{msftvectorsearchdocs}. These results are added in the rightmost column in the graphs in figure \ref{fig:hist}, but we note that these RAG figures are merely meant to contextualize our fine-tuned models' performances compared to standard practice. In none of the document/dataset-scaling configurations do we outperform RAG with fine-tuning, but this was a nongoal for this study. We get SFT performance to be within 16\% (or better) of RAG performance for all scenarios (closing much of the gap), but as discussed more rigorously in section \ref{sec:limitations}, our hyperparameters are not optimized, and the training has significant room for improvement. Here, we were primarily interested in the scaling.

\subsection{Fact Coverage in Token Datasets}
\label{sec:tokenfactcov}
At the core of our task is training to learn new knowledge, which we might represent as a set of constituent facts. By scaling up the token counts of our datasets, we aim to train our models with more examples covering those target facts. Consider however some particular fact or set of facts that our given model may struggle with -- token-based question-answer pair generation is not guaranteed to cover those weak areas as we naively scale up. There may be differences in the effectiveness of token-scaling as well due to different documents being more or less dense in new facts than others.

To explore the above cases further, we directly generate the set of atomic facts contained in each of the documents then retroactively study how many synthesized question-answer pairs cover each of the atomic facts. Coverage was estimated by transforming the atomic facts and question-answer pairs into embeddings using OpenAI's text-embedding-ada-002 model \citep{msftaoaimodeldocs} and calculating cosine similarity. We allowed multilabel assignment to account for the case where a single question might cover two closely related facts. The threshold for assignment was swept over, and 0.945 was selected as the value for determining assignment. 

\begin{figure}
\centering
\includegraphics[width=0.51\textwidth, keepaspectratio]{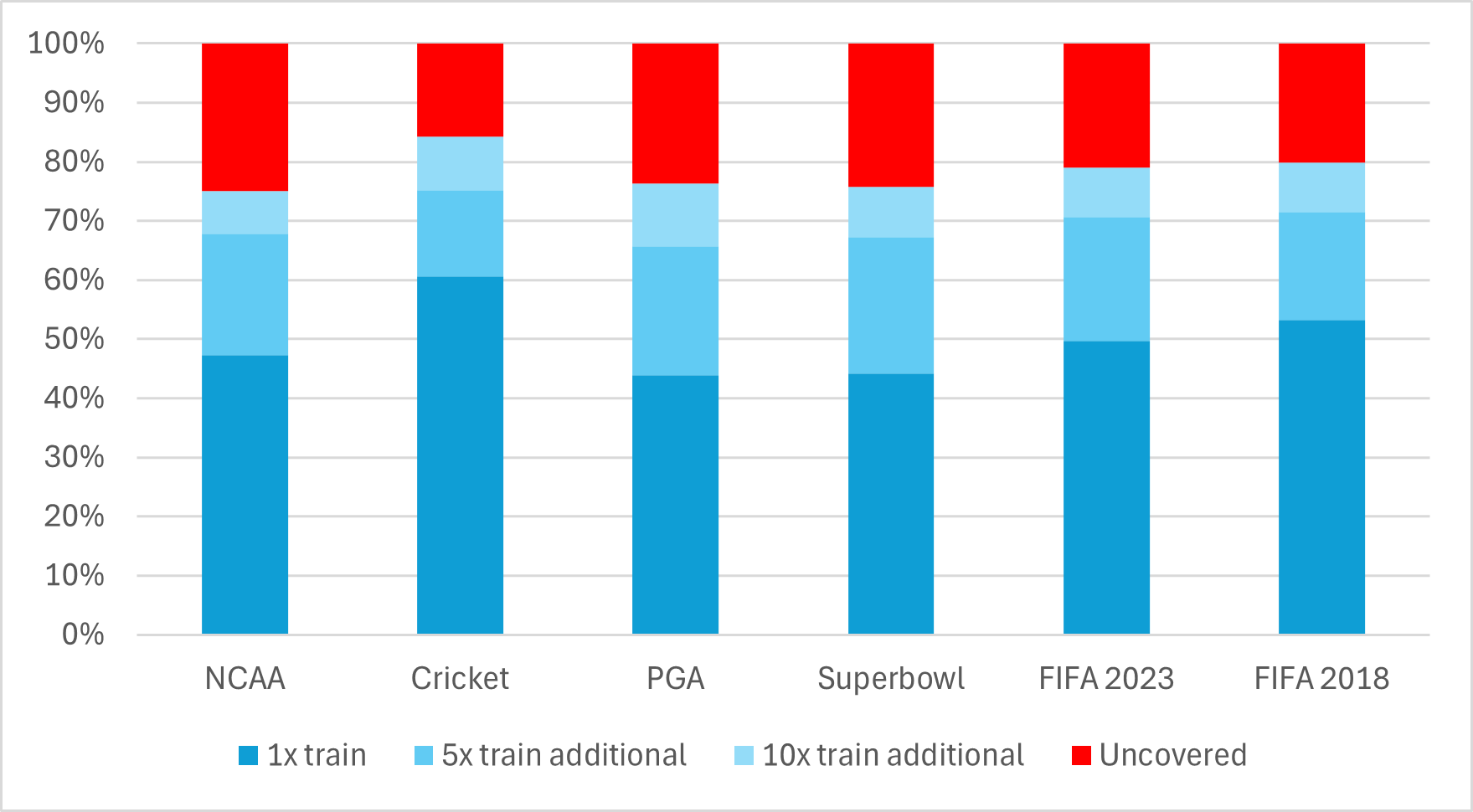}
\captionsetup{justification=centering}
    \setlength{\belowcaptionskip}{-8pt}
    \caption{Fact coverage across token-based datasets.}
    \label{fig:factcoverage}
\end{figure}

In our analysis, we would hope to see full coverage, meaning every atomic fact in a given document has at least one example in the training set that covers it. But as we can see in figure \ref{fig:factcoverage}, about 20\% of the facts are not covered by any example even in our 10x scaled datasets. For the facts that are covered, a desirable property might be an even distribution of question counts over them. We take to the FIFA 2023 Women's World Cup document and plot the number of related questions over all the atomic facts in the overview section in appendix  \ref{sec:tokfactdistribution}, finding this is not the case. Instead, we see some facts are far more popular than others in terms of question coverage, with one fact in the 10x scaling receiving more than 25 questions -- despite 20\% of the facts receiving none at all. 

The number of additional facts covered going from 1x to 5x is much greater across all our datasets than from 5x to 10x, offering another potential explanation for why our token-based training results showed such diminishing returns: the model may not be generating enough examples for its weakest areas as we scale. We note the token-based evaluation sets were generated according to the same procedure, however, meaning they have the same distribution, so such weak areas would likely fall in the low-coverage regime of facts rather than the no-coverage regime. That some facts may be covered by neither train nor evaluation set indicates a more serious gap in the methodology given our goal is to perform knowledge injection over the full extent of a document's content.

Clearly GPT-4 generated the questions for each section according to some distribution that is nonuniform; it is further unreasonable to assume this distribution will match that of a user querying about these events, so we are motivated to turn to an alternate dataset generation scheme in the hopes of covering more facts and increasing knowledge injection efficacy.

\subsection{Fact-Based Scaling}
\begin{figure*}[t]
\centering
\includegraphics[width=\textwidth, keepaspectratio]{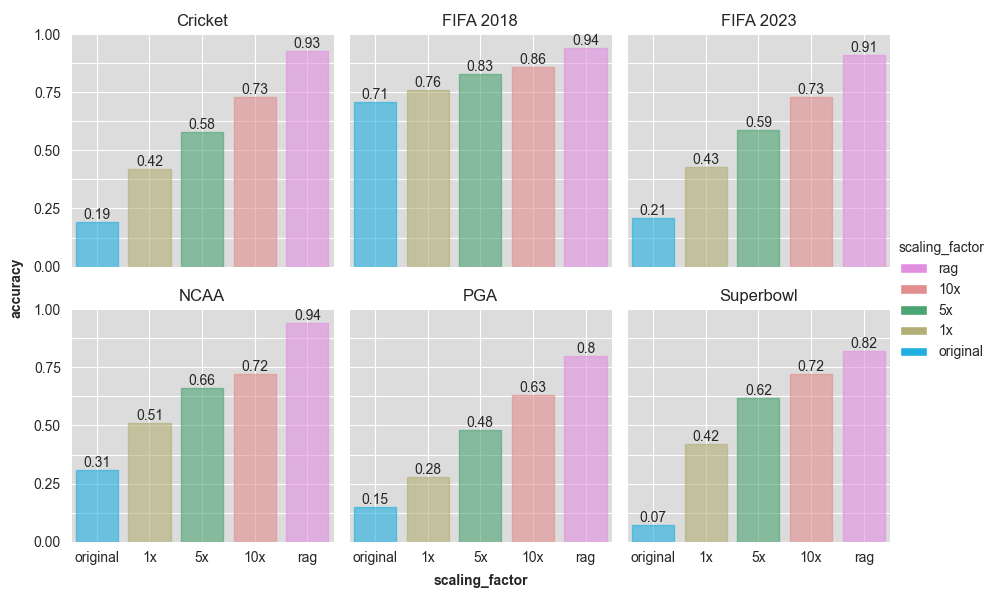}
    \caption{\textbf{Fact-based evaluation set} accuracy for our six documents across 1x, 5x, and 10x scaling with models trained on \textbf{fact-scaled datasets}. The base model results with no training are included under the bars annotated as original," and we include a RAG baseline as well which leverages the cleaned document sections to answer the eval questions.}
    \label{fig:factonfact}
\end{figure*}
In figure \ref{fig:factonfact}, we display the results from fine-tuning GPT-4 on datasets scaled over the fact dimension, scored on a fact-based evaluation set.

We make a few of the same observations as before: the base model does the best on FIFA 2018 at 0.712 vs. NCAA's 0.31, the best among the post-cutoff datasets. That NCAA jumped from 0.242 to 0.310 suggests some pre-cutoff facts may have not been covered in the previous token-based datasets. The second highest post-cutoff dataset is FIFA 2023 with a much lower base model accuracy of 0.211; once again we conclude the majority of facts from all post-cutoff datasets are out-of-domain.

Unlike with the token-based scaling, we see no negative drops in performance as we scale up, even for FIFA 2018. While again we do not beat RAG for any datasets, we notice scaling gains are more consistent as we go from 1x to 5x to 10x. We consider the FIFA 2018 base model performance a signal for knowledge ingestion efficacy after GPT-4's pretraining and observe the gap between our 10x trained models and RAG on the post-cutoff datasets is brought to parity to the gap between the FIFA 2018 base performance and RAG.

On the consistent scaling gains, the model is guaranteed to see additional relevant examples for facts it was struggling with as scaling increases, and performance goes up in turn. More importantly, as full and even fact coverage in the evaluation set is guaranteed according to our dataset generation procedure, we have more confidence in these results representing a complete picture of the model's efficacy at ingesting the full content of each document.

\subsection{Cross-Validating Token-on-Fact}
\label{sec:crossval}
\begin{figure*}
\centering
\includegraphics[width=\textwidth, keepaspectratio]{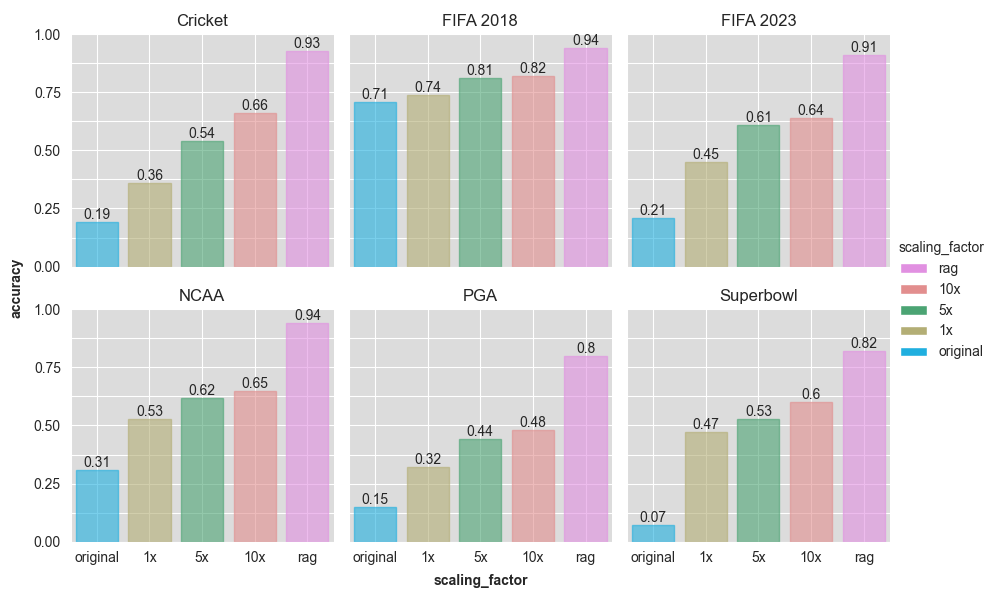}
    \caption{\textbf{Fact-based evaluation set} accuracy for our six documents across 1x, 5x, and 10x scaling with models trained on \textbf{token-scaled datasets}.}
    \label{fig:tokonfact}
\end{figure*}

With the more complete fact-based evaluation set, we go back and assess the models fine-tuned on the 1x, 5x, and 10x token-based scaling datasets. The Q\&A distributions are different due to the varying synthesis approaches, but the content and goals are the same, so we feel the comparison is especially valuable. The results are depicted in figure \ref{fig:tokonfact}.

The pre-cutoff FIFA 2018 results suffer the least degradation (up to a relative 10\%) compared to the post-cutoff datasets. The facts were already present in the base model from pretraining, so we rely on fine-tuning less for increased coverage. The post-cutoff datasets tell a different story; comparing the 1x results between figures \ref{fig:hist} and \ref{fig:tokonfact}, FIFA 2023 sees a small 5.5\% relative decrease from token (training)-on-token (eval) to token-on-fact; NCAA and Superbowl see small relative increases of about 2\% and 3\%, respectively; and PGA and Cricket see larger relative decreases on the order of 18-19\%. The significant differences from PGA, Cricket, and to a lesser extent FIFA 2023 speak to the uncovered facts in their 1x token datasets. We consider the small increases in the NCAA and Superbowl datasets as artifacts of randomness and lucky optimization, since in theory token-on-fact can only be as good as token-on-token due to the aforementioned coverage considerations. Once we hit the 5x scale, all models trained on token-scaled datasets see major degradations when evaluated on fact-based sets compared to token-based ones. The smallest such difference is FIFA 2023 which sees a relative decrease of -12.8\%, and the largest is PGA at -37.3\%. This trend is preserved at the 10x scale. For convenience, we plot out these relative degradations between figures \ref{fig:hist} and \ref{fig:tokonfact} in appendix \ref{sec:crossval_rels}.

Of additional interest is the difference in our scaling gains. The extent of these gains are lower across the board for token-on-fact compared to the token-on-token values from figure \ref{fig:hist}. This is attributed to what was described in section \ref{sec:tokenfactcov}: blindly scaling up based on token counts in the training sets does not gaurantee coverage of missed facts, particularly if GPT-4 is less likely to pay attention to them according to its distribution over Q\&A.

\section{Limitations and Future Work}
\label{sec:limitations}
This study uses the same sort of dataset for probing SFT and knowledge injection, Wikipedia articles about sporting championships. Future investigations could enrich the robustness of our findings by incorporating a variety of text corpora from different domains. This would not only validate the generalizability of our approach but also enhance the model's adaptability across a broader spectrum of knowledge domains. 

In assessing the efficacy of knowledge retention, we have focused on answer correctness within the context of sports-related content. However, it is essential to consider the model's performance on a wider range of topics, particularly in real-world scenarios where dialogue may traverse multiple domains. Future studies should examine the trade-offs involved in new knowledge acquisition, especially the potential impact on the model's proficiency in previously learned concepts.

\begin{figure}
    \includegraphics[width=\textwidth, keepaspectratio]{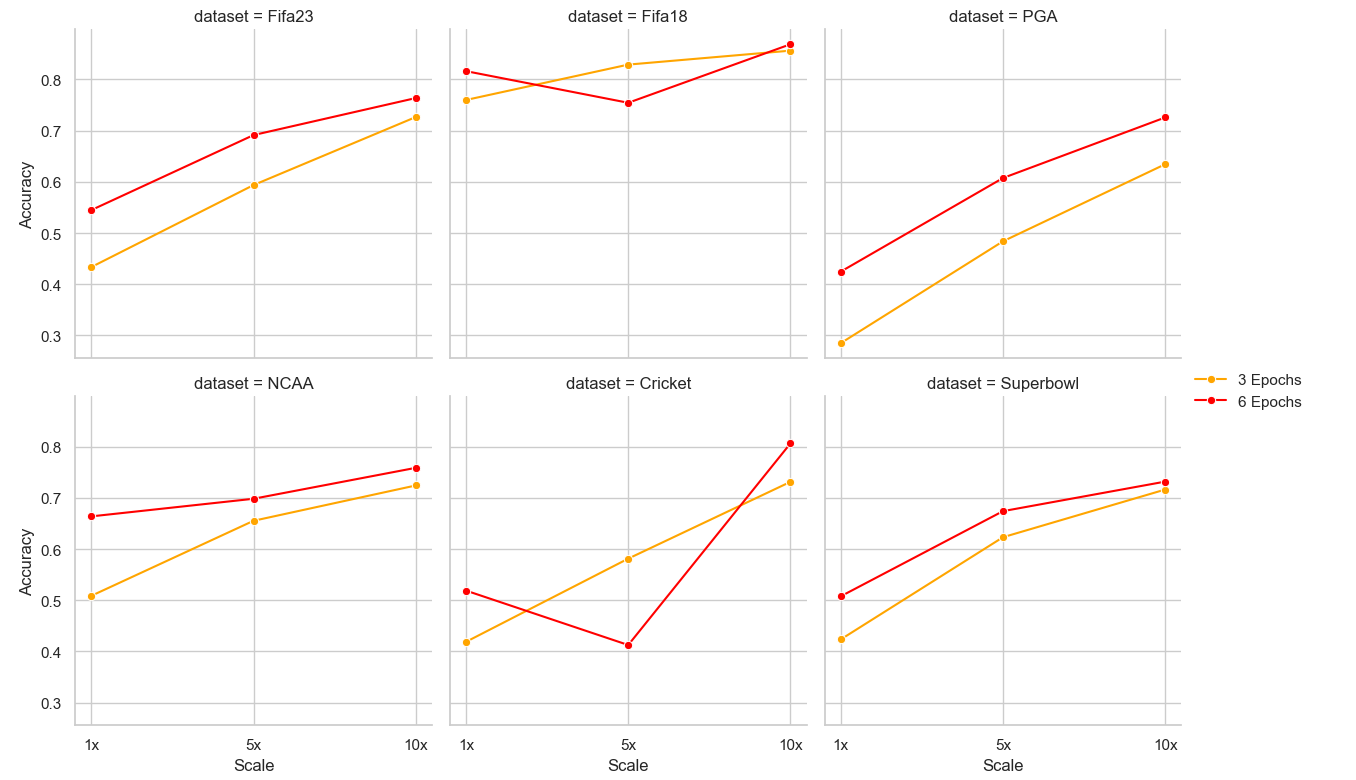}
    \caption{Fact scaling for 3- and 6- epochs. Note we see an anomaly in the Cricket 5x 6 epoch configuration where due to an unoptimized learning rate and our implementation which selects the last model checkpoint only, the end model falls into an extremely low training loss regime that leads to overfitting. In the random optimization over the 10x dataset, this does not occur.}
    \label{fig:epochcomp}
\end{figure}

Our preliminary exploration into the effects of hyperparameter tuning and the potential for underfitting has yielded promising results, with notable performance improvements observed upon increasing the number of training epochs. Here we try doubling the epochs from 3 to 6 and train on the fact datasets again, with results pictured in figure \ref{fig:epochcomp}. Some of our scores increase considerably on the evaluation set, as much as by an absolute 10\% at the 10x scale in the case of PGA or 15\% at the 1x scale in the case of NCAA. To what extent are the scaling gains we see merely a product of larger datasets equaling longer training, thus having the same effect as training for longer epochs? For this we look to compare the 5x 6 epoch results with the 10x 3 epoch ones, which should be roughly equal in terms of training budget but vary in terms of their scaling. In all six datasets, the 10x 3 epoch configuration outperforms the 5x 6 epoch one, though for some datasets this is only by 3-5\%. Some of our gains may therefore be attributed to simply longer training, but the consistent outperformance gives us confidence our scaling via different questions has its benefits too. Nonetheless, the relationship between dataset scaling, training duration, and learning efficacy merits further exploration. Future research should aim to delineate the specific contributions of dataset scaling from extended training periods, while also considering the risks of overfitting associated with increased epochs.

Moreover, the methodologies employed in generating our training and evaluation datasets, while consistent, may not fully encapsulate the diversity of language use encountered in real-world applications. As touched upon in section \ref{sec:tokscale}, our training and evaluation datasets were generated according to the same procedure and are coming from the same distribution. Future research could benefit from developing datasets that more accurately reflect the multifaceted nature of human inquiries, encompassing a range of linguistic styles and structures.

Finally, a direct comparison between models trained on token-scaled versus fact-scaled datasets presents certain complexities due to the variable nature of fact density and token counts. The token scaling does not have a fixed fact coverage due to the skewed Q\&A distribution of GPT-4, and the fact scaling does not have a fixed token count as the fact number itself may vary per document, and we need sufficient token budget to cover them all. We observe the fact-scaled datasets therefore have larger token counts than the token-scaled equivalents, with the 1x fact train sets having roughly 2x the tokens as the 1x token train sets. These fact-based experiments are therefore token-based scaling experiments in disguise, though we maintain our endorsement of the fact-based method due to the coverage guarantees discussed previously. Table \ref{table:facttokcounts} below lists the token-based multipliers for each fact-based train dataset. To address this, future work could aim to standardize one dimension—either tokens or facts—to facilitate clearer comparisons and draw more definitive conclusions about the relative merits of each approach.

\begin{table}
\caption{Token Multipliers for Fact-based Datasets}
\parbox{\textwidth}{
\vskip.15cm
\centerline{\begin{tabular}{ccccccc}
\hline  
 Scale Factor & NCAA & Cricket & PGA & Superbowl & FIFA 23 & FIFA 18 \\
\hline
1x & 1.87 & 2.66 & 2.12 & 2.02 & 2.17 & 2.04 \\
5x & 9.83 & 14.66 & 11.29 & 10.91 & 11.62 & 10.73 \\
10x & 19.74 & 28.69 & 22.67 & 21.77 & 23.35 & 21.49 \\
\end{tabular}  }}
\label{table:facttokcounts}
\end{table}

\clearpage
\bibliographystyle{plainnat}
\bibliography{paper}

\begin{thebibliography}{25}
\providecommand{\natexlab}[1]{#1}
\providecommand{\url}[1]{\texttt{#1}}
\expandafter\ifx\csname urlstyle\endcsname\relax
  \providecommand{\doi}[1]{doi: #1}\else
  \providecommand{\doi}{doi: \begingroup \urlstyle{rm}\Url}\fi

\bibitem[msf(2024{\natexlab{a}})]{msftaoaimodeldocs}
Azure openai service models, 2024{\natexlab{a}}.
\newblock URL \url{https://learn.microsoft.com/en-us/azure/ai-services/openai/concepts/models#gpt-4-and-gpt-4-turbo-preview}.
\newblock [Online; accessed 27-March-2024].

\bibitem[msf(2024{\natexlab{b}})]{msftvectorsearchdocs}
Azure openai on your data, 2024{\natexlab{b}}.
\newblock URL \url{https://learn.microsoft.com/en-us/azure/ai-services/openai/concepts/use-your-data?tabs=ai-search}.
\newblock [Online; accessed 27-March-2024].

\bibitem[Bai et~al.(2022)Bai, Kadavath, Kundu, Askell, Kernion, Jones, Chen, Goldie, Mirhoseini, McKinnon, Chen, Olsson, Olah, Hernandez, Drain, Ganguli, Li, Tran-Johnson, Perez, Kerr, Mueller, Ladish, Landau, Ndousse, Lukosuite, Lovitt, Sellitto, Elhage, Schiefer, Mercado, DasSarma, Lasenby, Larson, Ringer, Johnston, Kravec, Showk, Fort, Lanham, Telleen-Lawton, Conerly, Henighan, Hume, Bowman, Hatfield-Dodds, Mann, Amodei, Joseph, McCandlish, Brown, and Kaplan]{bai2022constitutional}
Yuntao Bai, Saurav Kadavath, Sandipan Kundu, Amanda Askell, Jackson Kernion, Andy Jones, Anna Chen, Anna Goldie, Azalia Mirhoseini, Cameron McKinnon, Carol Chen, Catherine Olsson, Christopher Olah, Danny Hernandez, Dawn Drain, Deep Ganguli, Dustin Li, Eli Tran-Johnson, Ethan Perez, Jamie Kerr, Jared Mueller, Jeffrey Ladish, Joshua Landau, Kamal Ndousse, Kamile Lukosuite, Liane Lovitt, Michael Sellitto, Nelson Elhage, Nicholas Schiefer, Noemi Mercado, Nova DasSarma, Robert Lasenby, Robin Larson, Sam Ringer, Scott Johnston, Shauna Kravec, Sheer~El Showk, Stanislav Fort, Tamera Lanham, Timothy Telleen-Lawton, Tom Conerly, Tom Henighan, Tristan Hume, Samuel~R. Bowman, Zac Hatfield-Dodds, Ben Mann, Dario Amodei, Nicholas Joseph, Sam McCandlish, Tom Brown, and Jared Kaplan.
\newblock Constitutional ai: Harmlessness from ai feedback, 2022.

\bibitem[Balaguer et~al.(2024)Balaguer, Benara, de~Freitas~Cunha, de~M.~Estevão~Filho, Hendry, Holstein, Marsman, Mecklenburg, Malvar, Nunes, Padilha, Sharp, Silva, Sharma, Aski, and Chandra]{balaguer2024rag}
Angels Balaguer, Vinamra Benara, Renato~Luiz de~Freitas~Cunha, Roberto de~M.~Estevão~Filho, Todd Hendry, Daniel Holstein, Jennifer Marsman, Nick Mecklenburg, Sara Malvar, Leonardo~O. Nunes, Rafael Padilha, Morris Sharp, Bruno Silva, Swati Sharma, Vijay Aski, and Ranveer Chandra.
\newblock Rag vs fine-tuning: Pipelines, tradeoffs, and a case study on agriculture, 2024.

\bibitem[Chen et~al.(2022)Chen, Zhang, Xie, Deng, Yao, Tan, Huang, Si, and Chen]{Chen_2022}
Xiang Chen, Ningyu Zhang, Xin Xie, Shumin Deng, Yunzhi Yao, Chuanqi Tan, Fei Huang, Luo Si, and Huajun Chen.
\newblock Knowprompt: Knowledge-aware prompt-tuning with synergistic optimization for relation extraction.
\newblock In \emph{Proceedings of the ACM Web Conference 2022}, WWW ’22. ACM, April 2022.
\newblock \doi{10.1145/3485447.3511998}.
\newblock URL \url{http://dx.doi.org/10.1145/3485447.3511998}.

\bibitem[Fan et~al.(2020)Fan, Gong, Wei, Wang, Huang, Jiao, Huang, Duan, and Zhang]{fan2020enhanced}
Zhihao Fan, Yeyun Gong, Zhongyu Wei, Siyuan Wang, Yameng Huang, Jian Jiao, Xuanjing Huang, Nan Duan, and Ruofei Zhang.
\newblock An enhanced knowledge injection model for commonsense generation.
\newblock \emph{arXiv preprint arXiv:2012.00366}, 2020.

\bibitem[Hu et~al.(2021)Hu, Shen, Wallis, Allen-Zhu, Li, Wang, Wang, and Chen]{hu2021lora}
Edward~J Hu, Yelong Shen, Phillip Wallis, Zeyuan Allen-Zhu, Yuanzhi Li, Shean Wang, Lu~Wang, and Weizhu Chen.
\newblock Lora: Low-rank adaptation of large language models.
\newblock \emph{arXiv preprint arXiv:2106.09685}, 2021.

\bibitem[Lauscher et~al.(2020)Lauscher, Majewska, Ribeiro, Gurevych, Rozanov, and Glava{\v{s}}]{lauscher2020common}
Anne Lauscher, Olga Majewska, Leonardo~FR Ribeiro, Iryna Gurevych, Nikolai Rozanov, and Goran Glava{\v{s}}.
\newblock Common sense or world knowledge? investigating adapter-based knowledge injection into pretrained transformers.
\newblock \emph{arXiv preprint arXiv:2005.11787}, 2020.

\bibitem[Lewis et~al.(2020)Lewis, Perez, Piktus, Petroni, Karpukhin, Goyal, K\"{u}ttler, Lewis, Yih, Rockt\"{a}schel, Riedel, and Kiela]{NEURIPS2020_6b493230}
Patrick Lewis, Ethan Perez, Aleksandra Piktus, Fabio Petroni, Vladimir Karpukhin, Naman Goyal, Heinrich K\"{u}ttler, Mike Lewis, Wen-tau Yih, Tim Rockt\"{a}schel, Sebastian Riedel, and Douwe Kiela.
\newblock Retrieval-augmented generation for knowledge-intensive nlp tasks.
\newblock In H.~Larochelle, M.~Ranzato, R.~Hadsell, M.F. Balcan, and H.~Lin, editors, \emph{Advances in Neural Information Processing Systems}, volume~33, pages 9459--9474. Curran Associates, Inc., 2020.
\newblock URL \url{https://proceedings.neurips.cc/paper_files/paper/2020/file/6b493230205f780e1bc26945df7481e5-Paper.pdf}.

\bibitem[Martino et~al.(2023)Martino, Iannelli, and Truong]{martino2023knowledge}
Ariana Martino, Michael Iannelli, and Coleen Truong.
\newblock Knowledge injection to counter large language model (llm) hallucination.
\newblock In \emph{European Semantic Web Conference}, pages 182--185. Springer, 2023.

\bibitem[McCloskey and Cohen(1989)]{MCCLOSKEY1989109}
Michael McCloskey and Neal~J. Cohen.
\newblock Catastrophic interference in connectionist networks: The sequential learning problem.
\newblock volume~24 of \emph{Psychology of Learning and Motivation}, pages 109--165. Academic Press, 1989.
\newblock \doi{https://doi.org/10.1016/S0079-7421(08)60536-8}.
\newblock URL \url{https://www.sciencedirect.com/science/article/pii/S0079742108605368}.

\bibitem[OpenAI(2023)]{openai2024gpt4}
OpenAI.
\newblock Gpt-4 technical report.
\newblock \emph{arXiv preprint arXiv:2303.08774}, 2023.
\newblock URL \url{https://doi.org/10.48550/arXiv.2303.08774}.

\bibitem[Ovadia et~al.(2024)Ovadia, Brief, Mishaeli, and Elisha]{ovadia2024finetuning}
Oded Ovadia, Menachem Brief, Moshik Mishaeli, and Oren Elisha.
\newblock Fine-tuning or retrieval? comparing knowledge injection in llms, 2024.

\bibitem[Rafailov et~al.(2023)Rafailov, Sharma, Mitchell, Ermon, Manning, and Finn]{rafailov2023direct}
Rafael Rafailov, Archit Sharma, Eric Mitchell, Stefano Ermon, Christopher~D. Manning, and Chelsea Finn.
\newblock Direct preference optimization: Your language model is secretly a reward model, 2023.

\bibitem[Schulman et~al.(2017)Schulman, Wolski, Dhariwal, Radford, and Klimov]{schulman2017proximal}
John Schulman, Filip Wolski, Prafulla Dhariwal, Alec Radford, and Oleg Klimov.
\newblock Proximal policy optimization algorithms, 2017.

\bibitem[Song et~al.(2016)Song, Yan, Li, Zhao, and Zhang]{song2016better}
Yiping Song, Rui Yan, Xiang Li, Dongyan Zhao, and Ming Zhang.
\newblock Two are better than one: An ensemble of retrieval- and generation-based dialog systems, 2016.

\bibitem[Tian et~al.(2023)Tian, Mitchell, Yao, Manning, and Finn]{tian2023finetuning}
Katherine Tian, Eric Mitchell, Huaxiu Yao, Christopher~D. Manning, and Chelsea Finn.
\newblock Fine-tuning language models for factuality, 2023.

\bibitem[Wang et~al.(2020)Wang, Tang, Duan, Wei, Huang, ji, Cao, Jiang, and Zhou]{wang2020kadapter}
Ruize Wang, Duyu Tang, Nan Duan, Zhongyu Wei, Xuanjing Huang, Jianshu ji, Guihong Cao, Daxin Jiang, and Ming Zhou.
\newblock K-adapter: Infusing knowledge into pre-trained models with adapters, 2020.

\bibitem[{Wikipedia contributors}(2024{\natexlab{a}})]{enwiki:1213024529}
{Wikipedia contributors}.
\newblock 2023 masters tournament --- {Wikipedia}{,} the free encyclopedia, 2024{\natexlab{a}}.
\newblock URL \url{https://en.wikipedia.org/w/index.php?title=2023_Masters_Tournament&oldid=1213024529}.
\newblock [Online; accessed 27-March-2024].

\bibitem[{Wikipedia contributors}(2024{\natexlab{b}})]{enwiki:1213361100}
{Wikipedia contributors}.
\newblock 2023 cricket world cup --- {Wikipedia}{,} the free encyclopedia, 2024{\natexlab{b}}.
\newblock URL \url{https://en.wikipedia.org/w/index.php?title=2023_Cricket_World_Cup&oldid=1213361100}.
\newblock [Online; accessed 27-March-2024].

\bibitem[{Wikipedia contributors}(2024{\natexlab{c}})]{enwiki:1214162630}
{Wikipedia contributors}.
\newblock 2023 fifa women's world cup --- {Wikipedia}{,} the free encyclopedia, 2024{\natexlab{c}}.
\newblock URL \url{https://en.wikipedia.org/w/index.php?title=2023_FIFA_Women%27s_World_Cup&oldid=1214162630}.
\newblock [Online; accessed 27-March-2024].

\bibitem[{Wikipedia contributors}(2024{\natexlab{d}})]{enwiki:1215541265}
{Wikipedia contributors}.
\newblock Super bowl lvii --- {Wikipedia}{,} the free encyclopedia, 2024{\natexlab{d}}.
\newblock URL \url{https://en.wikipedia.org/w/index.php?title=Super_Bowl_LVII&oldid=1215541265}.
\newblock [Online; accessed 27-March-2024].

\bibitem[{Wikipedia contributors}(2024{\natexlab{e}})]{enwiki:1215798602}
{Wikipedia contributors}.
\newblock 2023 ncaa division i men's basketball tournament --- {Wikipedia}{,} the free encyclopedia, 2024{\natexlab{e}}.
\newblock URL \url{https://en.wikipedia.org/w/index.php?title=2023_NCAA_Division_I_men%27s_basketball_tournament&oldid=1215798602}.
\newblock [Online; accessed 27-March-2024].

\bibitem[{Wikipedia contributors}(2024{\natexlab{f}})]{enwiki:1216022372}
{Wikipedia contributors}.
\newblock 2018 fifa world cup --- {Wikipedia}{,} the free encyclopedia, 2024{\natexlab{f}}.
\newblock URL \url{https://en.wikipedia.org/w/index.php?title=2018_FIFA_World_Cup&oldid=1216022372}.
\newblock [Online; accessed 27-March-2024].

\bibitem[Xu et~al.(2023)Xu, Namazifar, Hazarika, Padmakumar, Liu, and Hakkani-Tur]{xu-etal-2023-kilm}
Yan Xu, Mahdi Namazifar, Devamanyu Hazarika, Aishwarya Padmakumar, Yang Liu, and Dilek Hakkani-Tur.
\newblock {KILM}: Knowledge injection into encoder-decoder language models.
\newblock In Anna Rogers, Jordan Boyd-Graber, and Naoaki Okazaki, editors, \emph{Proceedings of the 61st Annual Meeting of the Association for Computational Linguistics (Volume 1: Long Papers)}, pages 5013--5035, Toronto, Canada, July 2023. Association for Computational Linguistics.
\newblock \doi{10.18653/v1/2023.acl-long.275}.
\newblock URL \url{https://aclanthology.org/2023.acl-long.275}.

\end{thebibliography}

\clearpage
\begin{appendices}
\section{Prompts}
\label{sec:sampleprompts}
Below is the prompt leveraged for token-based dataset generation. 
\newtcolorbox{myquote}{colback=block-gray,grow to right by=0mm,grow to left by=0mm,boxrule=0pt,boxsep=0pt,breakable}
\begin{myquote}
\textbf{System Prompt:} You are an AI assistant knowledgeable about recent sporting events. Be accurate but concise in response. \\
\textbf{User Prompt:} Please write 10 questions and answers probing the facts and statistics contained in the following snippet from an article about \{theme\}:\textbackslash n\textbackslash n\{passage\}\textbackslash n\textbackslash n1. Q: \{seed\_question\} A: \{question\_bank[seed\_question]\}\textbackslash n2. "
\end{myquote}

The following is the prompt used for atomic fact distillation. 
\begin{myquote}
\textbf{System Prompt:} You are an AI assistant knowledgeable about recent sporting events. Be accurate but concise in response. \\
\textbf{User Prompt:} Please break down the following snippet from an article about \{theme\} into atomic facts.\textbackslash nGoal 1: The atomic facts should be as simple as possible, if it's compound sentence, break down one more time.\textbackslash nGoal 2: For clarity, avoid using pronouns like 'it', 'he', 'she', 'this', 'that' etc., and instead use the full names or titles.\textbackslash n\textbackslash n\{passage\}\textbackslash n\textbackslash n1. 
\end{myquote}

Next is the prompt used for fact-based dataset generation with the skip option present.
\begin{myquote}
\textbf{System Prompt:} You are an AI assistant knowledgeable about recent sporting events. Be accurate but concise in response. \\
\textbf{User Prompt:} Write 10 pairs of questions and answers probing the facts and statistics of a given sentence about \{theme\}.\textbackslash nConsider first generating questions and answers that are very relevant and explicit to the fact, then paraphrase those questions and answers to reach the desired 10 Q\&A pairs. If the fact is too broad or not specific enough to \{theme\}, you may reply with ``SKIP" and be done.\textbackslash nEXAMPLE:\textbackslash nFACT: 14 million viewers tuned in to the opening game of the series.\textbackslash n1. Q: How many viewers watched the first game? A: 14 million people watched the first game of the series.\textbackslash n\textbackslash nEXAMPLE:\textbackslash nFACT: The rose is red.\textbackslash nSKIP\textbackslash n\textbackslash nFACT: \{fact\}\textbackslash n1. "
\end{myquote}

The fact-based dataset generation prompt with the skip option removed is the following:
\begin{myquote}
\textbf{System Prompt:} You are an AI assistant knowledgeable about recent sporting events. Be accurate but concise in response. \\
\textbf{User Prompt:} Write 10 pairs of questions and answers probing the fact and statistics of the below fact about \{theme\}:\textbackslash n\{fact\}\textbackslash n\textbackslash nYou can firstly generate questions and answers that are very relevant and explicit to the fact, and then paraphrase question and answers to reach the desired 10 pairs. \textbackslash n1. Q: sample question A: sample answer \textbackslash n2. 
\end{myquote}
In addition to removing the SKIP part, we found the altered phrasing here made the model more willing to generate Q\&A pairs on more general-sounding facts.

Finally, below is the prompt used for generated answer accuracy:
\begin{myquote}
\textbf{System Prompt:} You are a high school teacher grading student's responses for questions about {theme}. These responses are either correct or incorrect. \\
\textbf{User Prompt:} Please evaluate the correctness of a sentence in answering the question: ``\{question\}". The correct answer is ``\{ref\_answer\}". Your grading is binary: give 0 if the sentence is incorrect, give 1 if the sentence is correct. The sentence is ``\{pred\_answer\}." Please note that your output is either 0 or 1, no other information should be in the output.
\end{myquote}

The themes used in the above prompts are short snippets meant to contextualize each sporting event. For our six datasets, they are ``the 2023 Cricket World Cup", ``the 2018 FIFA World Cup", ``the recent 2023 FIFA Women's World Cup", ``the recent 2023 Superbowl LVII", ``the recent 2023 NCAA Division I Men's Basketball tournament", and ``the recent 2023 PGA Master's Tournament".

\clearpage
\section{FIFA 2023 Distribution of Facts over Token Datasets}
\label{sec:tokfactdistribution}
\begin{figure*}[hb!]
    \centering
    \begin{subfigure}{\textwidth}
        \centering
        \includegraphics[trim={0 0 0 1.5cm},clip,width=\textwidth]{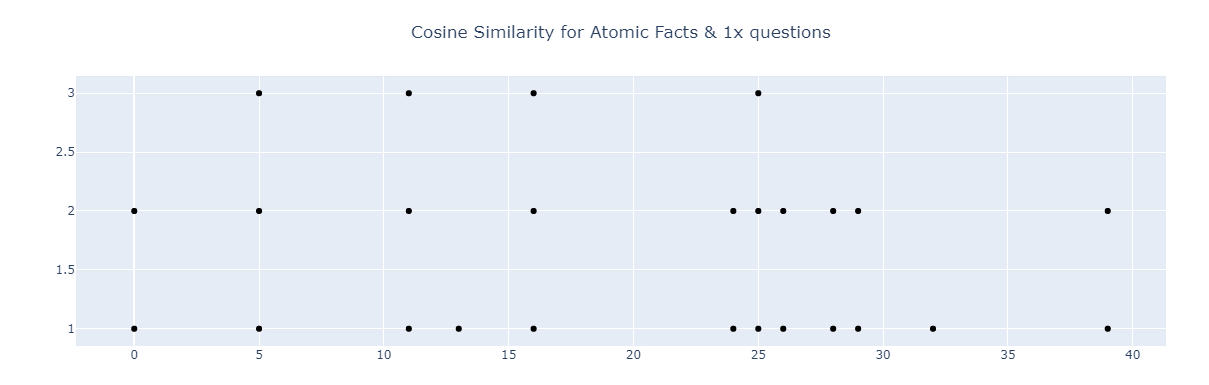}
        \caption{1x token dataset.}
    \end{subfigure}
    \begin{subfigure}{\textwidth}
        \centering
        \includegraphics[trim={0 0 0 1.5cm},clip,width=\textwidth]{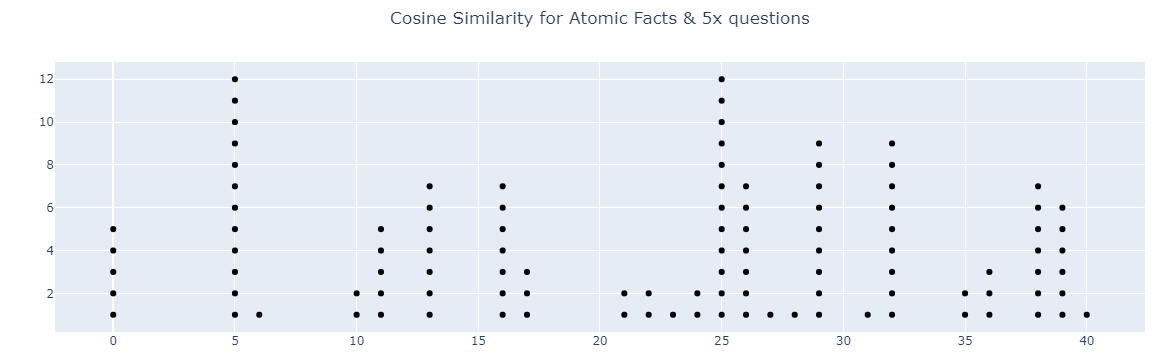}
        \caption{5x token dataset.}
    \end{subfigure}
    \begin{subfigure}{\textwidth}
        \centering
        \includegraphics[trim={0 0 0 1.5cm},clip,width=\textwidth]{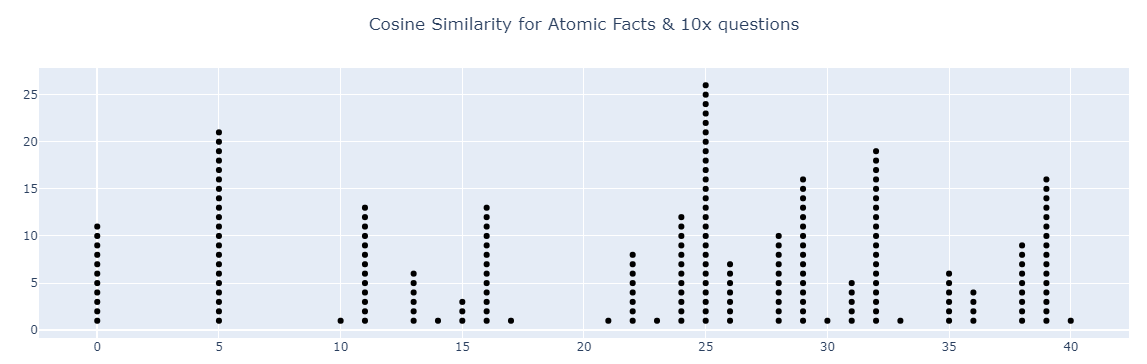}
        \caption{10x token dataset.}
    \end{subfigure}
    \caption{Number of related questions over all atomic facts in the overview section for FIFA 2023 dataset. Coverage is determined using cosine similarity.}
\end{figure*}

\clearpage
\section{Sample Atomic Facts}
\label{sec:atomicfacts}
Below we include 10 reference atomic facts extracted by GPT-4 for each of the six sporting events in this study. \\ \\
2023 Cricket World Cup: 
\begin{itemize}
    \itemsep0em
    \item Bangladesh qualified for the 2023 Cricket World Cup via the ICC Cricket World Cup Super League.
    \item Teams that did not progress past the league stage of the 2023 Cricket World Cup received \$100,000.
    \item Virat Kohli scored the most runs in the 2023 ICC Men's Cricket World Cup.
    \item Sean Easey is the head of the ICC's umpire selection panel.
    \item India had a 302-run win against Sri Lanka.
    \item Disney+ Hotstar televised matches in English and eight regional languages.
    \item Warm-up matches for the 2023 Cricket World Cup took place from 29 September to 3 October 2023.
    \item India was the first team to qualify for the semi-finals.
    \item The 2023 ICC Men's Cricket World Cup concluded on 19 November 2023.
    \item Sri Lanka was the only full member to progress from the knock-out qualification stage for the 2023 Cricket World Cup.
\end{itemize} 
2023 PGA Master's Tournament
\begin{itemize}
    \itemsep0em
    \item Brooks Koepka was leading by 4 strokes over Jon Rahm at the time of suspension.
    \item Patrick Cantlay had a round of 68 (4 under par).
    \item Jon Rahm won the 2023 Masters Tournament.
    \item Jon Rahm made three birdies and two bogeys on the back nine to finish at 134 (10 under par).
    \item Phil Mickelson was ten shots back in the 2023 Masters Tournament.
    \item Séamus Power became the fourth player in the event's history to make two holes-in-one in the same year.
    \item Viktor Hovland had no bogeys in his round.
    \item Three trees fell near the 17th tee moments before the second suspension.
    \item Jon Rahm was two under par for his round, three behind Brooks Koepka, when play was suspended on Friday.
    \item Play was suspended for the day at 3:16 pm Eastern time on Saturday, April 8, 2023.
\end{itemize}
2023 NCAA Division I Men's Basketball Tournament
\begin{itemize}
    \itemsep0em
    \item California State University, Sacramento was the host at the Golden 1 Center for the first and second rounds of the 2023 NCAA Division I Men's Basketball tournament.
    \item 32 automatic bids were filled by programs that won their conference tournament.
    \item 36 bids were issued ``at-large".
    \item Streaming was available through Paramount+.
    \item The defending national champions Kansas Jayhawks were eliminated in the second round of the 2023 NCAA Division I men's basketball tournament.
    \item The first and second rounds of the 2023 NCAA Division I Men's Basketball tournament were hosted at the Amway Center in Orlando, Florida on March 16 and 18.
    \item 16 consecutive tournaments since 2007 have seen a double-digit seed make the regional semifinals.
    \item Westwood One is a company.
    \item Three upsets occurred in the second round of the 2023 NCAA Division I Men's Basketball tournament.
    \item Texas Southern tied the Coppin State Eagles in 2008 and Liberty Flames in 2013 for most losses ever to make the tournament, with 20.
\end{itemize}
Here we make a few observations. The last sample fact in the list concerns pre-cutoff knowledge; indeed, the base model performed the best on the NCAA dataset of the five post-cutoff events, suggesting it had the highest frequency of pre-cutoff facts. Additionally, we can see the fact "Westwood One is a company" is too nonspecific to be new knowledge from the document, meaning this got past our prompt "SKIP" filtering. Spot-checking reveals only a handful of such examples across all six datasets, and we wonder if including a more explicit classification step for filtering post fact candidate generation might be more effective than our previous approach.
\newpage
2023 Superbowl LVII
\begin{itemize}
    \itemsep0em
    \item The game was also referred to as the ``Kelce Bowl".
    \item American country singer Chris Stapleton sang the national anthem at Superbowl LVII.
    \item The Eagles allowed 344 points in the season.
    \item The Super Bowl was the final for Norma Hunt, the widow of the Chiefs'
    \item The 2023 Super Bowl LVII marked the first time an Indigenous artist was commissioned to create the official artwork for the Super Bowl.
    \item The Eagles won the NFC Championship, 31–7, against the San Francisco 49ers.
    \item Patrick Mahomes won his second NFL Most Valuable Player award.
    \item Harrison Butker's game-winning kick was set up by a defensive holding call on Philadelphia cornerback James Bradberry.
    \item During this drive, Hurts completed two 17-yard passes to tight end Dallas Goedert.
\end{itemize}
2018 FIFA World Cup
\begin{itemize}
    \itemsep0em
    \item The location of the cities was chosen to reduce travel time for the teams.
    \item Goals scored from penalty shoot-outs in the 2018 FIFA World Cup are not counted towards an individual player's goal count.
    \item Preference was given to applicants with volunteering experience.
    \item The mascot was selected through a design competition among university students.
    \item Competing countries in the 2018 FIFA World Cup were divided into eight groups of four teams.
    \item Four matches of the 2018 FIFA World Cup were hosted in Yekaterinburg.
    \item The 2018 FIFA World Cup was the most expensive World Cup in history.
    \item Boris Johnson compared the event to the 1936 Olympics held in Nazi Germany.
    \item The Portugal/Spain bid came second in the voting for the 2018 FIFA World Cup host.
    \item Peru last appeared in 1982.
\end{itemize}
2023 FIFA Women's World Cup
\begin{itemize}
    \itemsep0em
    \item Zambia became the first landlocked country in Africa to qualify for a World Cup for either sex.
    \item Record numbers of fans watched the Matildas' games in the 2023 tournament.
    \item The visuals of Oceaunz nodded to the vast mountains of New Zealand.
    \item The draw started with pot one and ended with pot four.
    \item The opening match of the Women's World Cup took place in Auckland, New Zealand, on 20 July.
    \item The Matildas' performance increased public interest in football in Australia.
    \item The tournament replicated the format of the men's FIFA World Cup used between 1998 and 2022.
    \item The 2023 FIFA Women's World Cup prize pool was \$80 million greater than the previous tournament's prize pool.
    \item Works Creative Agency is based in Los Angeles.
    \item The Pakistani football association was suspended by FIFA.
\end{itemize}

\clearpage
\section{Cross-Validation Relative Differences}
\label{sec:crossval_rels}
\begin{figure*}[ht!]
\includegraphics[width=\textwidth, keepaspectratio]{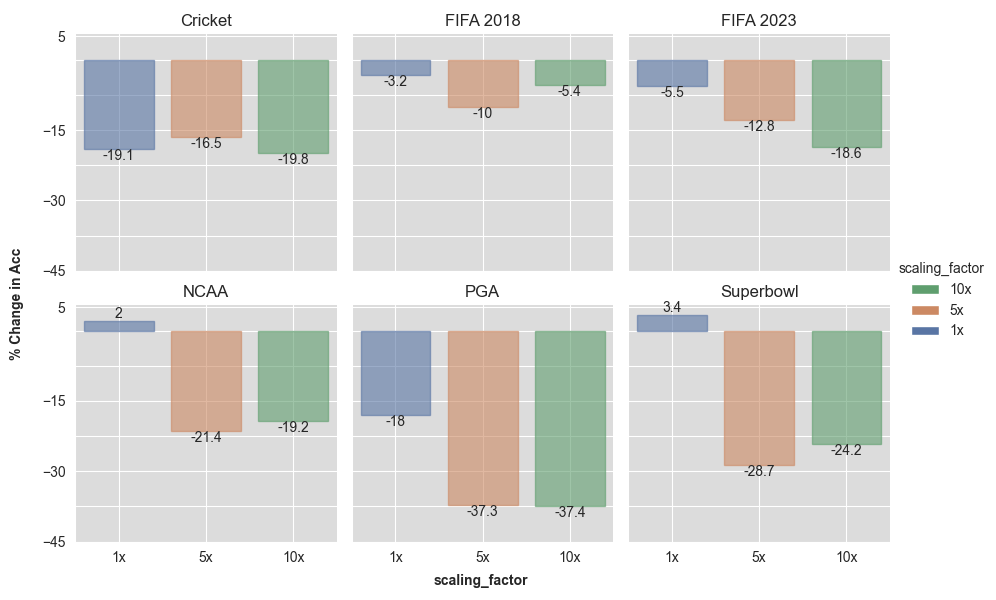}
\caption{Relative changes from evaluating the token-based models on token-based evaluation sets to fact-based evaluation sets, as discussed in section \ref{sec:crossval}.}
\end{figure*}

\end{appendices}

\end{document}